\documentclass{sig-alternate-2013}
\usepackage[pass]{geometry}

\newfont{\mycrnotice}{ptmr8t at 7pt}
\newfont{\myconfname}{ptmri8t at 7pt}
\let\crnotice\mycrnotice%
\let\confname\myconfname%

\permission{This is the authors' version of the work.  Copyright is
  held by the authors. The definitive version was published in ACM
  BuildSys'15, November 4--5, 2015,
  Seoul.\\ DOI:\href{http://dx.doi.org/10.1145/2821650.2821672}{10.1145/2821650.2821672}}

\conferenceinfo{}{
  {\mycrnotice{}}
  \copyrightetc{}
  \crdata{}
}

\usepackage[utf8]{inputenc}
\usepackage[usenames,dvipsnames]{color}
\usepackage{graphicx,url}
\usepackage{amsmath,xfrac} % maths packages
\usepackage{siunitx,booktabs}

\usepackage[
  bookmarks,pdfdisplaydoctitle,colorlinks, 
  urlcolor=BlueViolet, 
  citecolor=CadetBlue,
  linkcolor=black,
  pdfauthor={Jack Kelly},
  pdftitle={Neural NILM: Deep Neural Networks Applied to Energy
    Disaggregation}
]{hyperref}

\allowdisplaybreaks

\begin{document}

\title{Neural NILM: Deep Neural Networks\\Applied to Energy Disaggregation}

\numberofauthors{2}
\author{
  \alignauthor
  Jack Kelly\\
         \affaddr{Department of Computing}\\
         \affaddr{Imperial College London}\\
         \affaddr{180 Queen's Gate, London, SW7 2RH, UK}\\
         \email{jack.kelly@imperial.ac.uk}
  \alignauthor
  William Knottenbelt\\
         \affaddr{Department of Computing}\\
         \affaddr{Imperial College London}\\
         \affaddr{180 Queen's Gate, London, SW7 2RH, UK}\\  
         \email{w.knottenbelt@imperial.ac.uk}
}

\date{July 2015}

\maketitle
\begin{abstract}
Energy disaggregation estimates appliance-by-appliance electricity
consumption from a single meter that measures the whole home's
electricity demand. Recently, deep neural networks have driven
remarkable improvements in classification performance in neighbouring
machine learning fields such as image classification and automatic
speech recognition.  In this paper, we adapt three deep neural network
architectures to energy disaggregation: 1) a form of recurrent neural
network called `long short-term memory' (LSTM); 2) denoising
autoencoders; and 3) a network which regresses the start time, end
time and average power demand of each appliance activation.  We use
seven metrics to test the performance of these algorithms on real
aggregate power data from five appliances.  Tests are performed
against a house not seen during training and against houses seen
during training.  We find that all three neural nets achieve better F1
scores (averaged over all five appliances) than either combinatorial
optimisation or factorial hidden Markov models and that our neural net
algorithms generalise well to an unseen house.
\end{abstract}

%category{G.3}{Probability and Statistics}{Time series analysis}
% \category{I.5.1}{Pattern Recognition}{Models}[Neural nets]
%\category{I.5.4}{Pattern Recognition}{Applications}[Signal processing,
%waveform analysis]

%\terms{Experimentation, Algorithms, Design, Performance}

\section{Introduction}

Energy disaggregation (also called non-intrusive load monitoring or
NILM) is a computational technique for estimating the power demand of
individual appliances from a single meter which measures the combined
demand of multiple appliances.  One use-case is the production of
itemised electricity bills from a single, whole-home smart meter.  The
ultimate aim might be to help users reduce their energy consumption;
or to help operators to manage the grid; or to identify faulty
appliances; or to survey appliance usage behaviour.

Research on NILM started with the seminal work of
George Hart~\cite{hart1985, hart1992} in the mid-1980s.  Hart
described a `signature taxonomy' of features~\cite{hart1992} and his
earliest work from 1984 described experiments of extracting more
detailed features\footnote{This claim is taken from Hart
  1992~\cite{hart1992} because no copy of George Hart's 1984 technical
  report was available.}. However, Hart
decided to focus on extracting only transitions between steady-states.
Many NILM algorithms designed for low frequency data (1~Hz or slower)
follow Hart's lead and only extract a small number of features.  In
contract, in high frequency NILM (sampling at kHz or even MHz), there
are numerous examples in the literature of manually engineering rich
feature extractors (e.g. \cite{leeb1995transient,
  amirach2014feature-extraction}).

\begin{figure}
  \centering \epsfig{file=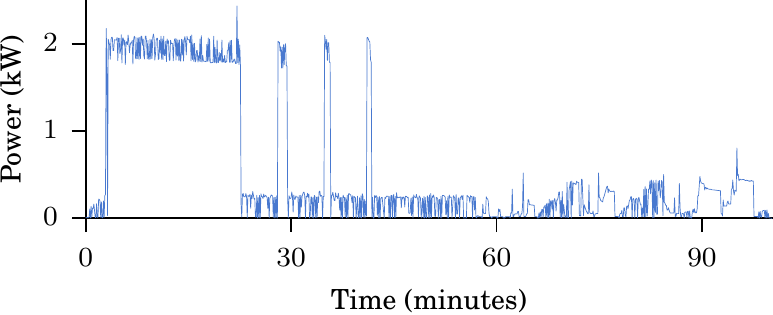}
  \caption{\label{fig:washer}Example power demand during one
    activation of the washing machine in UK-DALE House~1.}
\end{figure}

Humans can learn to detect appliances in aggregate data by eye,
especially appliances with feature-rich signatures such as the washing
machine signature shown in Figure~\ref{fig:washer}.  Humans almost
certainly make use of a variety of features such as the rapid on-off
cycling of the motor (which produces the rapid $\sim200$~watt
oscillations), the ramps towards the end as the washer starts to
rapidly spin the clothes etc.  We \textit{could} consider
hand-engineering feature extractors for these rich features.  But this
would be time consuming and the resulting feature detectors may not be
robust to noise and artefacts.  Two key research questions emerge:
Could an algorithm \textit{automatically learn} to detect these
features?  Can we learn anything from neighbouring machine learning
fields such as image classification?

Before 2012, the dominant approach to extracting features for image
classification was to hand-engineer feature detectors such as
scale-invariant feature transform~\cite{lowe1999SIFT} (SIFT) and
difference of Gaussians (DoG).  Then, in 2012, Krizhevsky \textit{et
  al.}'s winning algorithm~\cite{krizhevsky2012imagenet} in the
ImageNet Large Scale Visual Recognition Challenge achieved a
substantially lower error score (15\%) than the second-best approach
(26\%). Krizhevsky \textit{et al.}'s approach did not use
hand-engineered feature detectors.  Instead they used a deep neural
network which automatically learnt to extract a \textit{hierarchy of
  features} from the raw image.  Deep learning is now a dominant
approach not only in image classification but also fields such as
automatic speech
recognition~\cite{graves2014end-to-end-speech-recognition}, machine
translation~\cite{sutskever2014machine-translation}, even learning to
play computer games from
scratch~\cite{mnih2015deep-reinforcement-learning}!

In this paper, we investigate whether deep neural nets can be applied
to energy disaggregation.  The use of `small' neural nets on NILM
dates back at least to Roos \textit{et
  al.}\ 1994~\cite{roos1994neural-nets-nilm} (although that paper was
just a proposal) and continued with~\cite{yang2007neural-net-nilm,
  lin2010feature-extraction, ruzzelli2010,
  chang2011feature-extraction} but these small nets do not appear to
learn a hierarchy of feature detectors.  A big breakthrough in
image classification came when the compute power (courtesy of GPUs)
became available to train \textit{deep} neural networks on large
amounts of data. In the present research, we want to see if deep
neural nets can deliver good performance on energy disaggregation.

Our main contribution is to adapt three deep neural network
architectures to NILM.  For each architecture, we train one network
per target appliance.  We compare two benchmark disaggregation
algorithms (combinatorial optimisation and factorial hidden Markov
models) to the disaggregation performance of our three deep neural nets
using seven metrics.  We also examine how well our neural nets
generalise to appliances in houses not seen during training because,
ultimately, when NILM is used `in the field' we very rarely have
ground truth appliance data for the houses for which we want to
disaggregate.  So it is essential that NILM algorithms can generalise
to unseen houses.

Please note that, once trained, our neural nets \textit{do not} need
ground truth appliance data from each house!  End-users would only
need to provide aggregate data.  This is because each neural network
should learn the `essence' of its target appliance such that it can
generalise to unseen instances of that appliance.  In a similar
fashion, neural networks trained to do image classification are
trained on many examples of each category (dogs, cats, etc.) and
generalise to unseen examples of each category.

To provide more context, we will briefly sketch how our neural
networks could be deployed at scale, in the wild. Each net would
undergo \textit{supervised} training on \textit{many} examples of its
target appliance type so each network learns to generalise well to
unseen appliances.

Training is computationally expensive (days of processing on a fast
GPU).  But training does not have to be performed often. Once these
networks are trained, inference is much cheaper (around a second of
processing per network on a fast GPU for a week of aggregate data).
Aggregate data from unseen houses would be fed through each network.
Each network should filter out the power demand for its target
appliance.  This processing would probably be too computationally
expensive to run on an embedded processor inside a smart meter or
in-home-display.  Instead, the aggregate data could be sent from the
smart meter to the cloud.  The storage requirements for one 16~bit
integer sample (0-64~kW in 1~watt steps) every ten seconds is
17~kilobytes per day uncompressed.  This signal should be easily
compressible because there are numerous periods in domestic aggregate
power demand with little or no change.  With a compression ratio of
5:1, and ignoring the datetime index, the total storage
requirements for a year of data from 10~million users would be
13~terabytes (which could fit on two 8~TB disks). If one week of
aggregate data can be processed in one second per home (which should
be possible given further optimisation) then data from 10~million
users could be processed by 16 GPU compute nodes.  Alternatively,
disaggregation could be performed on a compute device within each home
(a modern laptop or mobile phone or a dedicated `disaggregation hub'
could handle the disaggregation).  A GPU is not \textit{required} for
disaggregation, although it makes it faster.

This paper is structured as follows: In Section~\ref{sec:IntroNets} we
provide a very brief introduction to artificial neural nets.  In
Section~\ref{sec:Data} we describe how we prepare the training data
for our nets and how we `augment' the training data by synthesising
additional data.  In Section~\ref{sec:Nets} we describe how we adapted
three neural net architectures to NILM.  In
Section~\ref{sec:Disaggregation} we describe how we do disaggregation
with our nets.  In Section~\ref{sec:Results} we present the
disaggregation results of our three neural nets and two benchmark NILM
algorithms.  Finally, in Section~\ref{sec:Conclusions} discuss our
results, offer our conclusions and describe some possible future
directions for research.

%% \begin{figure}
%%   \centering
%%   \includegraphics[width=\columnwidth]{harts_taxonomy.png}
%%   \caption{George Hart's `signature taxonomy'.  Taken from~\cite{hart1992}.}
%%   \label{fig:taxonomy}
%% \end{figure}

\section{Introduction to Neural Nets}
\label{sec:IntroNets}

An artificial neural network (ANN) is a directed graph where the nodes
are artificial neurons and the edges allow information from one neuron
to pass to another neuron (or the same neuron in a future time step).
Neurons are typically arranged into layers such that each neuron in
layer $l$ connects to every neuron in layer $l+1$.  Connections are
weighted and it is through modification of these weights that ANNs
learn.  ANNs have an \textit{input layer} and an
\textit{output layer}.  Any layers in between are called
\textit{hidden layers}. The \textit{forward pass} of an ANN is where
information flows from the input layer, through any hidden layers, to
the output.  Learning (updating the weights) happens during the
\textit{backwards pass}.

\subsection{Forwards pass}

Each artificial neuron calculates a weighted sum of its inputs, adds a
learnt bias and passes this sum through an activation function.
Consider a neuron which receives $I$ inputs.  The value of each input
is represented by input vector $x$.  The weight on the connection from
input $i$ to neuron $h$ is denoted by $w_{ih}$ (so $w$ is the `weights
matrix').  The weighted sum (also called the `network input') of the
inputs into neuron $h$ can be written $a_h = \sum_{i=1}^I x_iw_{ih}$.
The network input $a_h$ is then passed through an activation function
$\theta$ to produce the neuron's final output $b_h$ where $b_h =
\theta(a_h)$.  In this paper, we use the following activation
functions: linear: $\theta(x) = x$; rectified linear (ReLU):
$\theta(x) = \max(0, x)$; hyperbolic tangent (tanh): $\theta(x) =
\frac{\sinh x}{\cosh x} = \frac{e^x - e^{-x}}{e^x + e^{-x}}$.

Multiple nonlinear hidden layers can be used to re-represent the input
data (hopefully by learning a hierarchy of feature detectors), which
gives deep nonlinear networks a great deal of expressive
power~\cite{bengio2007scaling, hinton2006fast}.

\subsection{Backwards pass}

The basic idea of the backwards pass it to first do a forwards pass
through the entire network to get the network's output for a specific
network input.  Then compute the error of the output relative to the
target (in all our experiments we use the mean squared error (MSE) as
the objective function). Then modify the weights in the direction
which should reduce the error.

In practice, the forward pass is often computed over a \textit{batch}
of randomly selected input vectors.  In our work, we use a batch size
of 64 sequences per batch for all but the largest recurrent neural
network (RNN) experiments.  In our largest RNNs we use a batch size of
16 (to allow the network to fit into the 3GB of RAM on our GPU).

How do we modify each weight to reduce the error?  It would be
computationally intractable to enumerate the entire error surface.
MSE gives a smooth error surface and the activation functions are
differentiable hence we can use gradient descent.  The first step is
to compute the gradient of the error surface at the position for
current batch by calculating the derivative of the objective function
with respect to each weight.  Then we modify each weight by adding the
gradient multiplied by a `learning rate' scalar parameter.  To
efficiently compute the gradient (in $O(W)$ time) we use the
backpropagation algorithm~\cite{rumelhart1985learning,
  williams1995gradient, werbos1988generalization}.  In all our
experiments we use stochastic gradient descent (SGD) with Nesterov
momentum of $0.9$.

\subsection{Convolutional neural nets}

Consider the task of identifying objects in a photograph.  No matter
if we hand engineer feature detectors or learn feature detectors from
the data, it turns out that useful `low level' features concern small
patches of the image and include features such as edges of different
orientations, corners, blobs etc.  To extract these features, we want
to build a small number of feature detectors (one for horizontal
lines, one for blobs etc.) with small receptive fields (overlapping
sub-regions of the input image) and slide these feature detectors
across the entire image.  Convolutional neural
nets (CNNs)~\cite{fukushima1980neocognitron, atlas1988artificial,
  lecun1998gradient} build a small number of filters, each with
a small receptive field, and these filters are duplicated (with shared
weights) across the entire input.

Similarly to computer vision tasks, in time series problems we often
want to extract a small number of low level features with a small
receptive fields across the entire input.  All of our nets use at
least one 1D convolutional layer at the input.

\section{Training Data}
\label{sec:Data}
Deep neural nets need a lot of training data because they have a large
number of trainable parameters (the network weights and biases).  The
nets described in this paper have between 1~million to 150~million
trainable parameters.  Large training datasets are important.  It is
also common practice in deep learning to increase the effective size
of the training set by duplicating the training data many times and
applying realistic transformations to each copy.  For example, in
image classification, we might flip the image horizontally or apply
slight affine transformations.

A related approach to creating a large training dataset is to generate
simulated data.  For example, Google DeepMind train their
algorithms~\cite{mnih2015deep-reinforcement-learning} on computer
games because they can generate an effectively infinite amount of
training data.  Realistic synthetic speech audio data or natural
images are harder to produce.

In energy disaggregation, we have the advantage that generating
effectively infinite amounts of synthetic aggregate data is relatively
easy by randomly combining real appliance activations.  (We define an
`appliance activation' to be the power drawn by a single appliance
over one complete cycle of that appliance. For example,
Figure~\ref{fig:washer} shows a single activation for a washing
machine.)  We trained our nets on both synthetic aggregate data and
real aggregate data in a 50:50 ratio.  We found that synthetic data
acts as a regulariser.  In other words, training on a mix of synthetic
and real aggregate data rather than just real data appears to improve
the net's ability to generalise to unseen houses. For validation and
testing we use only real data (not synthetic).

We used UK-DALE~\cite{UK-DALE} as our source dataset.  Each submeter
in UK-DALE samples once every 6~seconds.  All houses record aggregate
apparent mains power once every 6~seconds.  Houses~1, 2 and 5 also
record active and reactive mains power once a second.  In these
houses, we downsampled the 1~second active mains power to 6~seconds to
align with the submetered data and used this as the real aggregate
data from these houses.  Any gaps in appliance data shorter than
3~minutes are assumed to be due to RF issues and so are filled by
forward-filling.  Any gaps longer than 3~minutes are assumed to be due
to the appliance and meter being switched off and so are filled with
zeros.

We manually checked a random selection of appliance activations from
every house.  The UK-DALE metadata shows that House~4's microwave and
washing machine share a single meter (a fact that we manually
verified) and hence these appliances from House~4 are not used in our
training data.

We train one network per target appliance.  The target (i.e. the
desired output of the net) is the power demand of the target
appliance.  The input to every net we describe in this paper is a
window of aggregate power demand.  The window width is decided on an
appliance-by-appliance basis and varies from 128 samples (13~minutes)
for the kettle to 1536 samples (2.5~hours) for the dish washer.  We
found that increasing the window size hurts disaggregation
performance for short-duration appliances (for example, using a
sequence length of 1024 for the fridge resulted in the autoencoder
(AE) failing to learn anything useful and the `rectangles' net
achieved an F1 score of 0.68; reducing the sequence length to 512
allowed the AE to get an F1 score of 0.87 and the `rectangles' net got
a score of 0.82).  On the other hand, it is important to ensure
that the window width is long enough to capture the majority of the
appliance activations.

For each house, we reserved the last week of data for testing and used
the rest of the data for training.  The number of appliance training
activations is show in Table~\ref{table:training_activations} and the
number of testing activations is shown in
Table~\ref{table:testing_activations}.  The specific houses used for
training and testing is shown in Table~\ref{table:houses}.

\subsection{Choice of appliances}

We used five target appliances in all our experiments: the fridge,
washing machine, dish washer, kettle and microwave.  We chose these
appliances because each is present in at least three houses in
UK-DALE.  This means that, for each appliance, we can train our nets
on at least two houses and test on a different house.  These five
appliances consume a significant proportion of energy and the five
appliances represent a range of different power `signatures' from the
simple on/off of the kettle to the complex pattern shown by the
washing machine (Figure~\ref{fig:washer}).

`Small' appliances such as games consoles and phone chargers are
problematic for many NILM algorithms because the effect of small
appliances on aggregate power demand tends to get lost in the noise.
By definition, small appliances do not consume much energy
individually but modern homes tend to have a large number of such
appliances so their combined consumption can be significant.  Hence it
would be useful to detect small appliances using NILM. We have not
explored whether our neural nets perform well on `small' appliances
but we plan to in the future.

\subsection{Extract activations}

Appliance activations are extracted using
NILMTK's~\cite{NILMTK}\\ \texttt{Electric.get\_activations()} method.
The arguments we passed to \texttt{get\_activations()} for each
appliance are shown in Table~\ref{table:get_activations_arguments}.
On simple appliances such as toasters, we extract activations by
finding strictly consecutive samples above some threshold power.  We
then throw away any activations shorter than some threshold duration
(to ignore spurious spikes).  For more complex appliances such as
washing machines whose power demand can drop below threshold for short
periods during a cycle, NILMTK ignores short periods of sub-threshold
power demand.

\subsection{Select windows of real aggregate data}

First we locate all the activations of the target appliance in the
home's submeter data for the target appliance.  Then, for each
training example, the code decides with 50\% probability whether this
example should include the target appliance or not.  If the code
decides not include the target appliance then it finds a random window
of aggregate data in which there are no activations of the target
appliance.  Otherwise, the code randomly selects a target appliance
activation and randomly positions this activation within the window of
data that will be shown to the net as the target (with the constraint
that the activation must be captured completely in the window of data
shown to the net, unless the window is too short to contain the entire
activation).  The corresponding time window of real aggregate data is
also loaded and shown to the net and its input.  If other activations
of the target appliance happen to appear in the aggregate data then
these are not included in the target sequence; the net is trained to
focus on the first complete target appliance activation in the
aggregate data.

\subsection{Synthetic aggregate data}

To create synthetic aggregate data we start by extracting a set of
appliance activations for five appliances across all training houses:
kettle, washing machine, dish washer, microwave and fridge.  To create
a single sequence of synthetic data, we start with two vectors of
zeros: one vector will become the input to the net; the other will
become the target.  The length of each vector defines the `window
width' of data that the network sees.  We go through the five
appliance classes and decide whether or not to add an activation of
that class to the training sequence.  There is a 50\% chance that the
target appliance will appear in the sequence and a 25\% chance for
each other `distractor' appliance.  For each selected appliance class,
we randomly select an appliance activation and then randomly pick
where to add that activation on the input vector.  Distractor
appliances can appear anywhere in the sequence (even if this means
that only part of the activation will be included in the sequence).
The target appliance activation must be completely contained within
the sequence (unless it is too large to fit).

Of course, this relatively naïve approach to synthesising aggregate
data ignores a lot of structure that appears in real aggregate data.
For example, the kettle and toaster might often appear within a few
minutes of each other in real data, but our simple `simulator' is
completely unaware of this sort of structure.  We expect that a more
realistic simulator might increase the performance of deep neural nets
on energy disaggregation.

\subsection{Implementation of data processing}

All our code is written in Python and we make use Pandas, Numpy and
NILMTK for data preparation.  Each network receives data in a
mini-batch of 64 sequences (except for the large RNN sequences, in
which case we use a batch size of 16 sequences).  The code is
multi-threaded so the CPU can be busy preparing one batch of data on
the fly whilst the GPU is busy training on the previous batch.

\subsection{Standardisation}

In general, neural nets learn most efficiently if the input data has
zero mean. First, the mean of each sequence is subtracted from the
sequence to give each sequence a mean of zero.  Every input sequence
is divided by the standard deviation of a random sample of the
training set.  We do not divide each sequence by its \textit{own}
standard deviation because that would change the scaling and the
scaling is likely to be important for NILM.

Forcing each sequence to have zero mean throws away information.
Information that NILM algorithms such as combinatorial optimisation
and factorial hidden Markov models rely on.  We have done some
preliminary experiments and found that neural nets appear to be able to
generalise better if we independently centre each sequence.  But there
are likely to be ways to have the best of both worlds: i.e. to give
the network information about the absolute power whilst also allowing
the network to generalise well.

One big advantage of training our nets on sequences which have been
independently centred is that our nets do not need to consider vampire
(always on) loads.

Targets are divided by a hand-coded `maximum power demand' for each
appliance to put the target power demand into the range [0, 1].

\begin{table}
\caption{Number of training activations per house.}
\label{table:training_activations}
\centering  
\begin{tabular}{ l  S[table-format=5]  S[table-format=4]
    S[table-format=2]  S[table-format=4]  S[table-format=4] }
  \toprule
   & {1} & {2} & {3} & {4} & {5} \\
  \midrule
  Kettle & 2836 & 543 & 44 & 716 & 176 \\
  Fridge & 16336 & 3526 & 0 & 4681 & 1488 \\
  Washing machine & 530 & 53 & 0 & 0 & 51 \\
  Microwave & 3266 & 387 & 0 & 0 & 28 \\
  Dish washer & 197 & 98 & 0 & 23 & 0 \\
  \bottomrule
\end{tabular}
\end{table}

\begin{table}
\caption{Number of testing activations per house.}
\label{table:testing_activations}
\centering  
\begin{tabular}{ l  S[table-format=3]  S[table-format=3]
    S[table-format=2]  S[table-format=3]  S[table-format=3] }
  \toprule
   & {1} & {2} & {3} & {4} & {5} \\
  \midrule
  Kettle &   54 &  29 & 40 &  50 &  18 \\
  Fridge &   168 &  277 & 0 &  145 &  140 \\
  Washing machine &  10 & 4 & 0 & 0 &  2 \\
  Microwave &   90 &   9 & 0 & 0 &  4 \\
  Dish washer &   3 &  7 & 0 &  3 \\
  \bottomrule
\end{tabular}
\end{table}

\begin{table}
\caption{Houses used for training and testing.}
\label{table:houses}
\centering
\begin{tabular}{ l  l  c }
  \toprule
   & Training & Testing \\
  \midrule
  Kettle & 1, 2, 3, 4 & 5 \\
  Fridge &  1, 2, 4 & 5 \\
  Washing machine & 1, 5 & 2 \\
  Microwave & 1, 2 & 5 \\
  Dish washer & 1, 2 & 5 \\
  \bottomrule
\end{tabular}
\end{table}

\begin{table}
  \caption{Arguments passed to \texttt{get\_activations()}.}
  \label{table:get_activations_arguments}
  \centering
\begin{tabular}{ l  S[table-format=4]  S[table-format=4]
    S[table-format=4]  S[table-format=4]}
  \toprule
            & {Max}   & {On power}  & {Min. on}  & {Min. off} \\
  Appliance & {power} & {threshold} & {duration} & {duration} \\
            & {(watts)}   & {(watts)}       & {(secs)}    & {(secs)}    \\
  \midrule
  Kettle & 3100 & 2000 & 12 & 0 \\
  Fridge & 300 & 50 & 60 & 12 \\
  Washing m. & 2500 & 20 & 1800 & 160\\
  Microwave & 3000 & 200 & 12 & 30 \\
  Dish washer & 2500 & 10 & 1800 & 1800 \\
  \bottomrule
\end{tabular}
\end{table}

\section{Neural Network Architectures}
\label{sec:Nets}

In this section we describe how we adapted three different neural net
architectures to do NILM.

\subsection{Recurrent Neural Networks}

In Section~\ref{sec:IntroNets} we described \textit{feed forward}
neural networks which map from a single input vector to a
single output vector.  When the network is shown a second input
vector, it has no memory of the previous input.

Recurrent neural networks (RNNs) allow cycles in the network graph
such that the output from neuron $i$ in layer $l$ at time step $t$ is
fed via weighted connections to every neuron in layer $l$ (including
neuron $i$) at time step $t + 1$.  This allows RNNs, in principal, to
map from the \textit{entire history} of the inputs to an output
vector.  This makes RNNs especially well suited to sequential data.
In our work, we train RNNs using backpropagation through time
(BPTT)~\cite{werbos1990BPTT}.

In practice, RNNs can suffer from the `vanishing gradient'
problem~\cite{hochreiter1997LSTM} where gradient information
disappears or explodes as it is propagated back through time.  This
can limit an RNN's memory.  One solution to this problem is the `long
short-term memory' (LSTM) architecture~\cite{hochreiter1997LSTM} which
uses a `memory cell' with a gated input, gated output and gated
feedback loop.  The intuition behind LSTM is that it is a
differentiable latch (where a `latch' is the fundamental unit of a
digital computer's RAM).  LSTMs have been used with success on a wide
variety of sequence tasks including automatic speech
recognition~\cite{graves2014end-to-end-speech-recognition,
  chorowski2014end-to-end-speech} and machine
translation~\cite{sutskever2014machine-translation}.

An additional enhancement to RNNs is to use \textit{bidirectional}
layers.  In a bidirectional RNN, there are effectively two parallel
RNNs, one reads the input sequence forwards and the other reads
the input sequence backwards.  The output from the forwards and backwards
halves of the network are combined either by concatenating them or
doing an element-wise sum (we experimented with both and settled on
concatenation, although element-wise sum appeared to work almost as
well and is computationally cheaper).

We should note that bidirectional RNNs are not naturally suited to
doing online disaggregation.  Bidirectional RNNs could still be used
for online disaggregation if we frame `online disaggregation' as doing
\textit{frequent, small batches} of offline disaggregation.

We experimented with both RNNs and LSTMs and settled on the
following architecture for energy disaggregation:

\begin{enumerate}
\itemsep0em
\item Input (length determined by appliance duration)
\item 1D conv (filter size=4, stride=1, number of filters=16,
  activation function=linear, border mode=same)
\item bidirectional LSTM (N=128, with peepholes)
\item bidirectional LSTM (N=256, with peepholes)
\item Fully connected (N=128, activation function=TanH)
\item Fully connected (N=1, activation function=linear)
\end{enumerate}

At each time step, the network sees a single sample of aggregate power
data and outputs a single sample of power data for the target appliance.

In principal, the convolutional layer should not be necessary (because
the LSTMs should be able to remember all the context).  But we found
the addition of a convolution layer to slightly increase
performance (the conv. layer convolves over the time axis).  We also
experimented with adding a conv. layer \textit{between} the two LSTM
layers with a stride >~1 to implement hierarchical
subsampling~\cite{graves2012book-supervised-sequence-labelling}.  This
showed promise but we did not use it for our final experiments.

On the backwards pass, we clip the gradient at
[-10, 10] as per Alex Graves in~\cite{graves2013generating-sequences}.
To speed up computation, we propagate the gradient backwards a maximum
of 500 time steps.  Figure~\ref{fig:net_output} shows an example
output of our LSTM network in the two `RNN' rows.

%% \begin{figure}
%% \centering
%% \epsfig{file=e566_kettle_rnn_train_estimates_10000epochs_8.pdf, width=\columnwidth} 
%% \caption{\label{fig:LSTMoutput}Example output of our bidirectional LSTM 
%%   trained on kettles.  This example is of the net doing
%%   disaggregation on aggregate data from a house not seen during testing.}
%% \end{figure}

\subsection{Denoising Autoencoders}
In this section, we frame energy disaggregation as a `denoising' task.
Typical denoising tasks include removing grain from an old photograph;
or removing reverb from an audio recording; or even in-filling a
masked part of an image.  Energy disaggregation can be viewed as an
attempt to recover the `clean' power demand signal of the target
appliance from the background `noise' produced by the other
appliances.  A successful neural network architecture for denoising
tasks is the `denoising autoencoder'.

An autoencoder (AE) is simply a network which tries to reconstruct the
input.  Described like this, AEs might not sound very useful!  The key
is that AEs first \textit{encode} the input to a \textit{compact}
vector representation (in the `code layer') and then \textit{decode}
to reconstruct the input.  The simplest way of forcing the net to
discover a \textit{compact} representation of the data is to have a
code layer with less dimensions than the input.  In this case,
the AE is doing dimensionality reduction.  Indeed, a linear AE with a
single hidden layer is almost equivalent to PCA.  But AEs can be deep
and non-linear.

A denoising autoencoder (dAE)~\cite{vincent2008denoising-autoencoders}
is an autoencoder which attempts to reconstruct a clean target from a
noisy input.  dAEs are typically trained by artificially corrupting a
signal before it goes into the net's input, and using the clean signal
as the net's target.  In NILM, we consider the corruption as being the
power demand from the other appliances.  So we do not add noise
artificially.  Instead we use the aggregate power demand as the
(noisy) input to the net and ask the net to reconstruct the clean
power demand of the target appliance.

The first and last layers of our NILM dAEs are 1D convolutional
layers.  We use convolutional layers because we want the network to
learn low level feature detectors which are applied equally across the
\textit{entire} input window (for example, a step change of 1000~watts
might be a useful feature to extract, no matter where it is found in
the input).  The aim is to provide some invariance to where exactly
the activation is positioned within the input window.  The last layer
does a `deconvolution'.

The exact architecture is as follows:

\begin{enumerate}
\itemsep0em  
\item Input (length determined by appliance duration)
\item 1D conv (filter size=4, stride=1, number of filters=8,
  activation function=linear, border mode=valid)
\item Fully connected (N=(sequence length - 3) $\times$ 8,\\
  activation function=ReLU)
\item Fully connected (N=128; activation function=ReLU)
\item Fully connected (N=(sequence length - 3) $\times$ 8,\\
  activation function=ReLU)  
\item 1D conv (filter size=4, stride=1, number of filters=1,
  activation function=linear, border mode=valid)
\end{enumerate}

Layer 4 is the middle, code layer. The entire dAE is trained
end-to-end in one go (we do not do layer-wise pre-training as we found
it did not increase performance).  We do not tie the weights as we
found this also appears to not enhance NILM performance.  An example output
of our NILM dAE is shown in Figure~\ref{fig:net_output} in the two
`Autoencoder' rows.

%% \begin{figure}
%% \centering
%% \epsfig{file=e567_washing_machine_ae_train_estimates_41788epochs_14.pdf, width=\columnwidth} 
%% \caption{\label{fig:dAEoutput}Example output of a denoising
%%   autoencoder trained on washing machines.  The bottom subplot is the
%%   input to the net: standardised, real aggregate power from a house in
%%   the training set. The middle plot is the ground truth target.  The
%%   top plot is the network's reconstruction of the washing machine.
%%   The X-axis scale is number of samples (at 6~second intervals).  The
%%   Y-axis represents power.}
%% \end{figure}

\subsection{Regress Start Time, End Time \& Power}
Many applications of energy disaggregation do not require a detailed
second-by-second reconstruction of the appliance power demand.
Instead, most energy disaggregation use-cases require, for each
appliance activation, the identification of the start time, end time
and energy consumed.  In other words, we want to draw a rectangle
around each appliance activation in the aggregate data where the left
side of the rectangle is the start time, the right side is the end
time and the height is the average power demand of the appliance
between the start and end times.

Deep neural networks have been used with great success on related
tasks.  For example, Nouri used deep neural networks to estimate the
2D location of `facial keypoints' in images of
faces~\cite{nouri2014facial-keypoints}.  Example `keypoints' are `left
eye centre' or `mouth centre top lip'.  The input to Nouri's neural
net is the raw image of a face.  The output of the network is a set of
$x,y$ coordinates for each keypoint.

Our idea was to train a neural network to estimate three scalar,
real-valued outputs: the start time, the end time and mean power
demand of the first appliance activation to appear in the aggregate
power signal. If there is no target appliance in the aggregate data
then all three outputs should be zero.  If there is more than one
activation in the aggregate signal then the network should ignore all
but the first activation.  All outputs are in the range [0, 1].  The
start and end times are encoded as a proportion of the input's time
window.  For example, the start of the time window is encoded as 0,
the end is encoded as 1 and half way through the time window is
encoded as 0.5.  For example, consider a scenario where the input
window width is 10~minutes and an appliance activation starts 1~minute
into the window and ends 1~minute before the end of the window.  This
activation would be encoded as having a start location of 0.1 and an
end location of 0.9.  Example output is shown in
Figure~\ref{fig:net_output} in the two `Rectangles' rows.

The three target values for each sequence are calculated during data
pre-processing.  As for all of our other networks, the network's
objective is to minimise the mean squared error. The exact
architecture is as follows:

\begin{enumerate}
\itemsep0em
\item Input (length determined by appliance duration)
\item 1D conv (filter size=4, stride=1, number of filters=16,
  activation function=linear, border mode=valid)
\item 1D conv (filter size=4, stride=1, number of filters=16,
  activation function=linear, border mode=valid)
\item Fully connected (N=4096, activation function=ReLU)
\item Fully connected (N=3072; activation function=ReLU)
\item Fully connected (N=2048, activation function=ReLU)
\item Fully connected (N=512, activation function=ReLU)
\item Fully connected (N=3, activation function=linear)
\end{enumerate}

%% \begin{figure}
%% \centering
%% \epsfig{file=disag_example.pdf, width=\columnwidth} 
%% \caption{\label{fig:rectangles_output}Example output of the net
%%   trained to regress the start time, end time and mean power of each
%%   appliance activation.  This example also shows the result of sliding
%%   the network over 50~hours of aggregate data (bottom subplot) to
%%   produce `overlapping rectangles' of estimated appliance activations
%%   (top plot) which are then thresholded to produce a vector output.}
%% \end{figure}

\subsection{Neural net implementation}

We implemented our neural nets in Python using the\\
\href{https://github.com/Lasagne/Lasagne}{Lasagne
  library}\footnote{\href{https://github.com/Lasagne/Lasagne}{github.com/Lasagne/Lasagne}}.
Lasagne is built on top of
\href{http://www.deeplearning.net/software/theano/}{Theano}~\cite{bergstra2010theano,
  bastien2012theano}. We trained our nets on an nVidia GTX 780Ti GPU
with 3~GB of RAM (but note that Theano also allows code to be run on
the CPU without requiring any changes to the user's code).  On this
GPU, our nets typically took between 1 and 12 hours to train per
appliance.  The exact code used to create the results in paper is
available in our
`\href{https://github.com/JackKelly/neuralnilm_prototype}{NeuralNILM
  Prototype}' repository\footnote{\href{https://github.com/JackKelly/neuralnilm_prototype}{github.com/JackKelly/neuralnilm\_prototype}}
and a more elegant (hopefully!) re-write is available in our
`\href{https://github.com/JackKelly/neuralnilm}{NeuralNILM}' repository\footnote{\href{https://github.com/JackKelly/neuralnilm}{github.com/JackKelly/neuralnilm}}.

We manually defined the number of weight updates to perform during
training for each experiment.  For the RNNs we performed 10,000
updates, for the denoising autoencoders we performed 100,000 and for the
regression network we performed 300,000 updates.  Neither the RNNs nor
the AEs appeared to continue learning past this number of updates. The
regression networks appear to keep learning no matter how many updates
we perform!

The nets have a wide variation in the number of trainable parameters.
The largest dAE nets range from 1M to 150M (depending on the input
size); the RNNs all had 1M parameters and the regression nets varied
from 28M to 120M parameters (depending on the input size).

All our network weights were initialised randomly using Lasagne's
default initialisation.  All of the experiments presented in this
paper trained end-to-end from random initialisation (no layerwise
pre-training).

\section{Disaggregation}
\label{sec:Disaggregation}
How do we disaggregate arbitrarily long sequences of aggregate data given
that each net has an input window duration of, at most, a few hours?
We first pad the beginning and end of the input with zeros.  Then we
slide the net along the input sequence.  As such, the first sequence
we show to the network will be all zeros.  Then we shift the input
window \texttt{STRIDE} samples to the right, where \texttt{STRIDE} is
a manually defined positive, non-zero integer.  If \texttt{STRIDE} is
less than the length of the net's input window then the net will see
overlapping input sequences.  This allows the network to have multiple
attempts at processing each appliance activation in the aggregate
signal, and on each attempt each activation will be shifted to the
left by \texttt{STRIDE} samples.

Over the course of disaggregation, the network produces multiple
estimated values for each time step because we give the network
overlapping segments of the input.  For our first two network
architectures, we combine the multiple values per timestep simply by
taking the mean.

Combing the output from our third network is a little more complex.
We layer every predicted `appliance rectangle' on top of each other.
We measure the overlap and normalise the overlap to [0, 1].  This
gives a probabilistic output for each appliance's power demand.  To
convert this to a single vector per appliance, we threshold the power
and probability.

\section{Results}
\label{sec:Results}
\begin{figure}
\centering
\epsfig{file=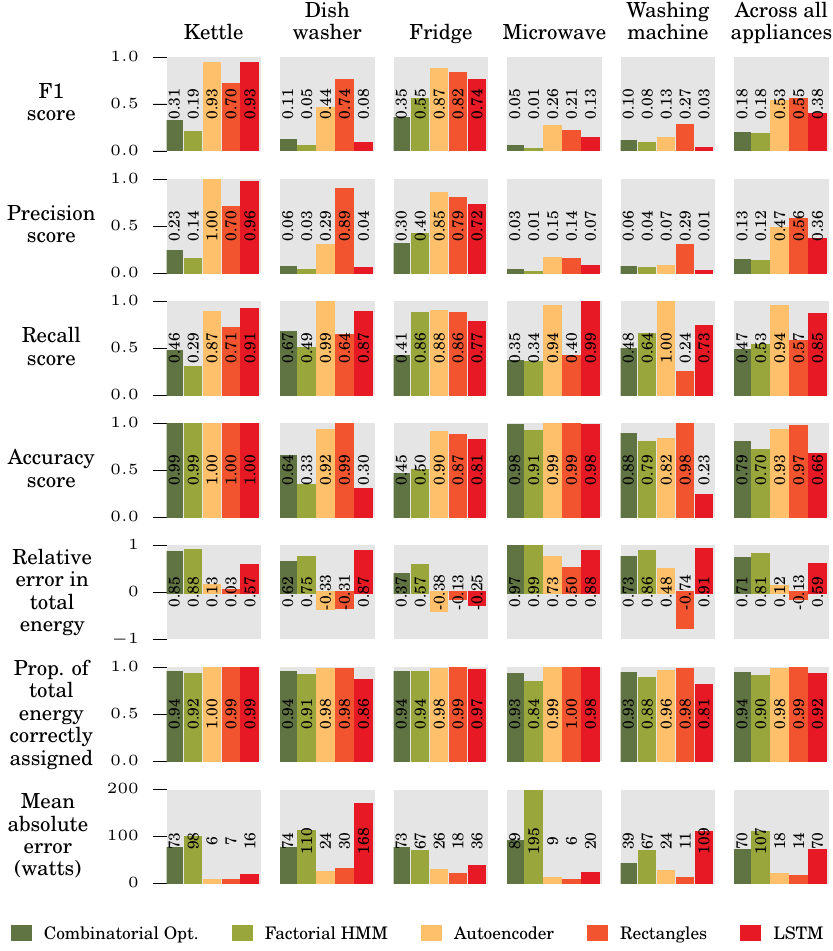, width=\columnwidth} 
\caption{\label{fig:results_unseen_house}Disaggregation performance on
  a house not seen during training.}
\end{figure}

\begin{figure}
\centering
\epsfig{file=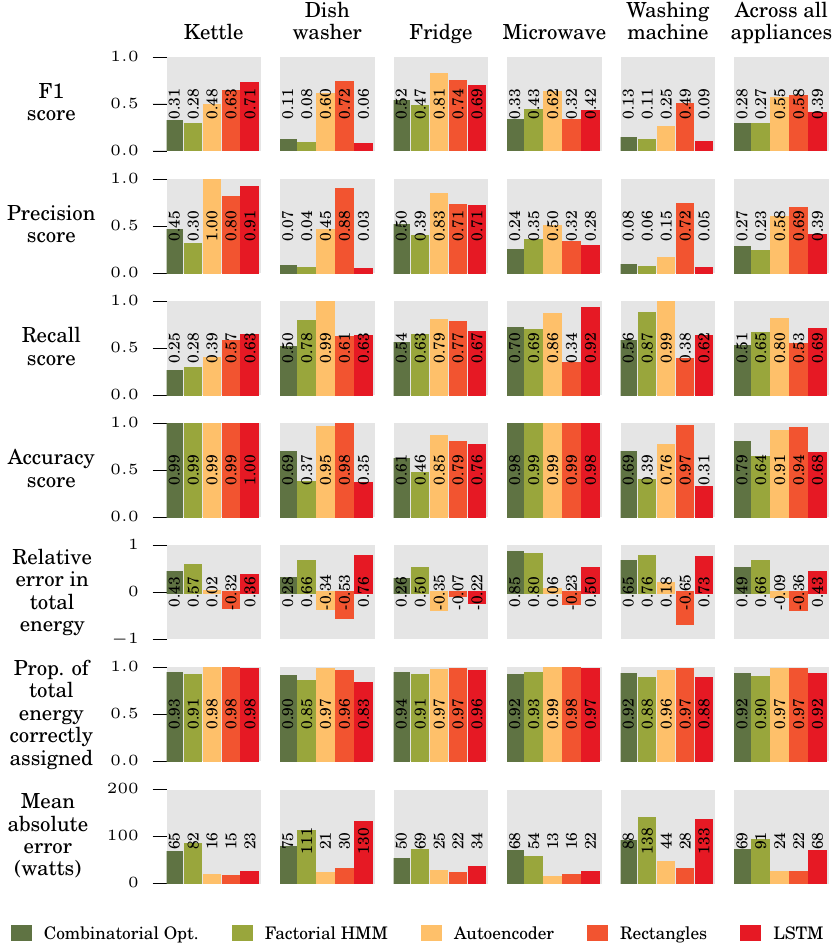, width=\columnwidth}
\caption{\label{fig:results_train_houses}Disaggregation performance on houses seen during training
  (the time window used for testing is different to that used
  for training).}
\end{figure}

The disaggregation results on an unseen house are shown in
Figure~\ref{fig:results_unseen_house}.  The results on houses seen
during training are shown in Figure~\ref{fig:results_train_houses}.

We used benchmark implementations from NILMTK~\cite{NILMTK} of the
combinatorial optimisation (CO) and factorial hidden Markov model (FHMM)
algorithms. 

On the unseen house (Figure~\ref{fig:results_unseen_house}), both the
denoising autoencoder and the net which regresses the start time, end
time and power demand (the `rectangles' architecture) out-perform CO
and FHMM on every appliance on F1 score, precision score, proportion
of total energy correctly assigned and mean absolute error.  The LSTM
out-performs CO and FHMM on two-state appliances (kettle, fridge and
microwave) but falls behind CO and FHMM on multi-state appliances
(dish washer and washing machine).

On the houses seen during training
(Figure~\ref{fig:results_train_houses}), the dAE outperforms CO and
FHMM on every appliance on every metric except relative error in total
energy.  The `rectangles' architecture outperforms CO and FHMM on
every appliance (except the microwave) on F1, precision, accuracy,
proportion of total energy correctly assigned and mean absolute error.

The full disaggregated time series for all our algorithms and the
aggregate data and appliance ground truth data are available at
 \href{http://www.doc.ic.ac.uk/~dk3810/neuralnilm}{www.doc.ic.ac.uk/$\sim$dk3810/neuralnilm}

The metrics we used are:

\begin{align}
  \textbf{TP} &= \text{number of true positives} \\
  \textbf{FP} &= \text{number of false positives} \\
  \textbf{FN} &= \text{number of false negatives} \\
  \textbf{P}  &= \text{number of positives in ground truth} \\
  \textbf{N}  &= \text{number of negatives in ground truth} \\
  \mathbf{E} &= \text{total actual energy} \\
  \mathbf{\hat{E}} &= \text{total predicted energy} \\
  \mathbf{y_t^{(i)}} &=
    \text{appliance } i \text{ actual power at time } t \\
  \mathbf{\hat{y}_t^{(i)}} &=
    \text{appliance } i \text{ estimated power at time } t \\
  \mathbf{\bar{y}_t} &=
    \text{aggregate actual power at time } t \\  
  \textbf{recall} &= \frac{\text{TP}}{\text{TP} + \text{FN}} \\
  \textbf{precision} &= \frac{\text{TP}}{\text{TP} + \text{FP}} \\
  \textbf{F1} &= 2 \times \frac
         {\text{precision} \times \text{recall}}
         {\text{precision} + \text{recall}} \\
  \textbf{accuracy} &= \frac
         {\text{TP} + \text{TN}}
         {\text{P} + \text{N}} \\
  \textbf{relative err} & \textbf{or in total energy} = \frac
         {|\hat{E} - E|}
         {\max(E, \hat{E})} \\
  \textbf{mean absol} & \textbf{ute error} = \sfrac{1}{T}
         \sum_{t=1}^T |\hat{y}_t - y_t|
\end{align}

\textbf{proportion of total energy correctly assigned} =
\begin{align}
  1 - \frac
       {\sum_{t=1}^{T} \sum_{i=1}^{n} | \hat{y}_t^{(i)} - y_t^{(i)} |}
       {2 \sum_{t=1}^T \bar{y}_t}
\end{align}

The proportion of total energy correctly assigned is taken
from~\cite{kolter2011REDD}. 

\begin{figure*}
  \centering \epsfig{file=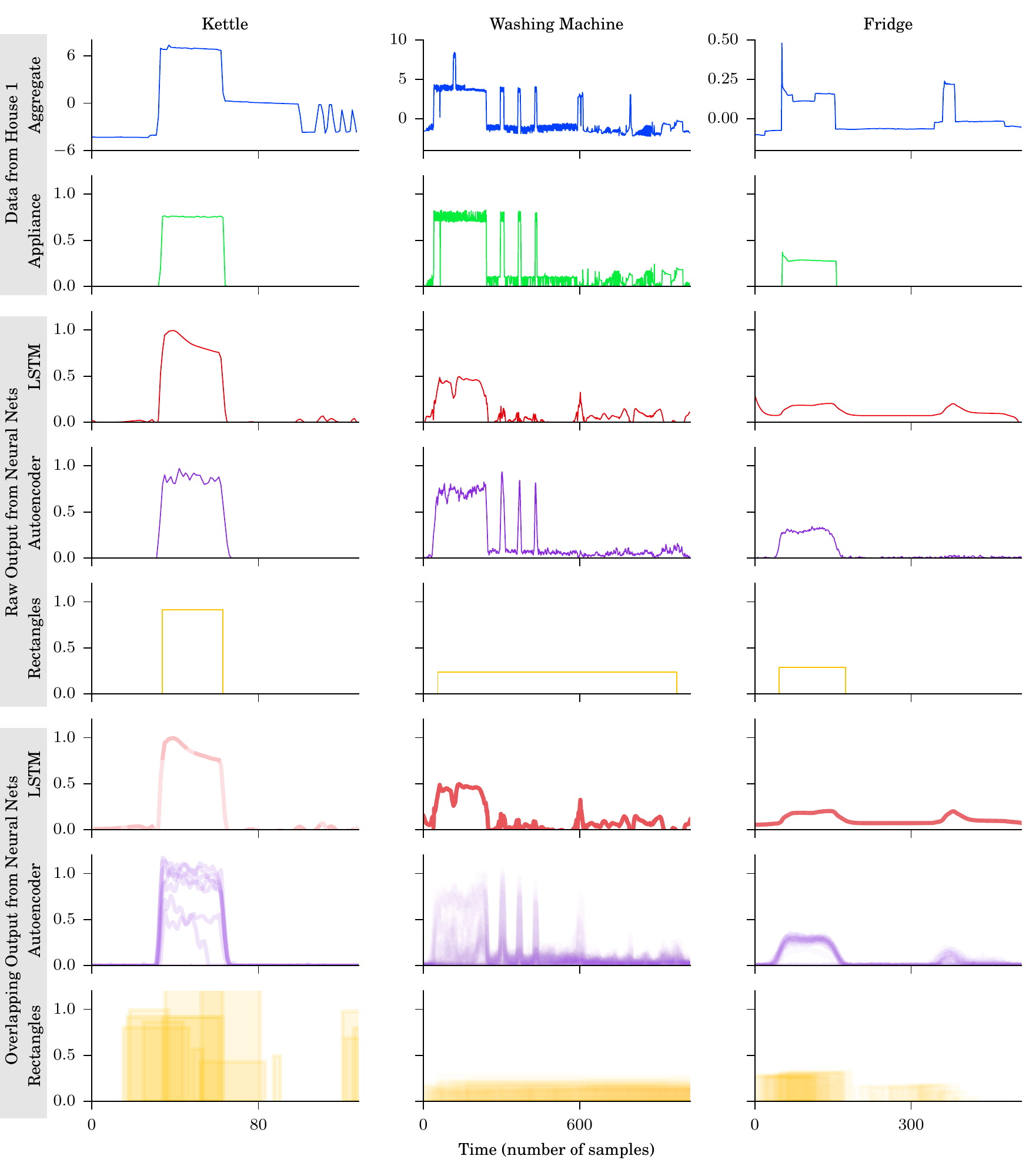}
  \caption{\label{fig:net_output} Example outputs produced by all
    three neural network architectures for three appliances.  Each
    column shows data for a different appliance.  The rows are in
    three groups (the tall grey rectangles on the far left).  The top
    group shows measured data from House~1.  The top row shows the
    measured aggregate power data from House~1 (the input to the
    neural nets).  The Y-axis scale for the aggregate data is
    standardised such that its mean is $0$ and its standard deviation
    is $1$ across the data set.  The Y-axis range for all other
    subplots is $[0, 1]$.  The second row shows the single-appliance
    power demand (i.e. what the neural nets are trying to estimate).
    The middle group of rows shows the raw output from each neural
    network (just a single pass through each network).  The bottom
    group of rows shows the result of sliding the network over the
    aggregate data with \texttt{STRIDE=16} and overlapping the output.
    Please note that the `rectangles' net is trained such that the
    height of the output rectangle should be the \textit{mean} power
    demand over the duration of the identified activation.}
\end{figure*}

\section{Conclusions \& Future Work}
\label{sec:Conclusions}
We have adapted three neural network architectures to NILM.  The
denoising autoencoder and the `rectangles' architectures perform well,
especially on unseen houses.  We believe that deep neural nets show
great promise for NILM.  But there is plenty of work still to do!

It is worth noting that our comparison between each architecture is
not entirely fair because the architectures have a wide range of
trainable parameters.  For example, every LSTM we used had 1M
parameters whilst the larger dAE and rectangles nets had over 150M
parameters (we did try training an LSTM with more parameters but it
did not appear to improve performance).

Our LSTM results suggest that LSTMs work best for two-state appliances
but do not perform well on multi-state appliances such as the dish
washer and washing machine.  One possible reason is that, for these
appliances, informative `events' in the power signal can be many time
steps apart (e.g. for the washing machine there might be over 1,000
time steps between the first heater activation and the spin cycle).
In principal, LSTMs have an arbitrarily long memory. But these long
gaps between informative events may present a challenge for LSTMs.
Further work is required to understand exactly why LSTMs struggle on
multi-state appliances.  One aspect of our LSTM results that we
\textit{did} expect was that processing overlapping windows of
aggregate data would not be necessary for LSTMs because they always
output the same estimates, no matter what the offset of the input
window (see Figure~\ref{fig:net_output}).

We must also note that the FHMM implementation used in this work is
not `state of the art' and neither is it especially tuned.  Other FHMM
implementations are likely to perform better.  We encourage other
researchers to download\footnote{Data available from
  \href{http://www.doc.ic.ac.uk/~dk3810/neuralnilm}{www.doc.ic.ac.uk/$\sim$dk3810/neuralnilm}}
our disaggregation estimates and ground truth data and directly
compare against our algorithms!

This work represents just a first step towards adapting the vast
number of techniques from the deep learning community to NILM, for
example:

\subsection{Train on more data}
UK-DALE has many hundreds of days of data but only from five houses.
Any machine learning algorithm is only able to generalise if given
enough variety in the training set.  For example, House~5's dish
washer sometimes has four activations of its heater but the dish
washers in the two training houses (1 and 2) only ever have two peaks.
Hence the autoencoder completely ignores the first two peaks of
House~5's dish washer!  If neural nets are to learn to generalise well
then we must train on much larger numbers of appliances (hundreds or
thousands).  This should help the networks to generalise across the
wide variation seen in some classes of appliance.

%% \begin{figure}
%% \centering
%% \epsfig{file=e567_dish_washer_ae_estimates_100000epochs_26.pdf, width=\columnwidth}
%% \caption{\label{fig:dish_washer}Disaggregation performance of the
%%   denoising autoencoder on a dish washer from an unseen house.}
%% \end{figure}

\vfill\eject
\subsection{Unsupervised pre-training}
In NILM, we generally have access to much more \textit{unlabelled}
data than \textit{labelled} data.  One advantage of neural nets is
that they could, in principal, be `pre-trained' on unlabelled data
before being fine-tuned on labelled data.  `Pre-training' should allow
the networks to start to identify useful features from the data but
does not allow the nets to learn to \textit{label} appliances.
(Pre-training is rarely used in modern image classification tasks
because very large labelled datasets are available for image
classification.  But in NILM we have much more unlabelled data than
labelled data, so pre-training is likely to be useful.)  After
unsupervised pre-training, each net would undergo \textit{supervised}
training.  Instead of (or as well as) pre-training on all available
unlabelled data, it may also be interesting to try pre-training
largely on unlabelled data from each house that we wish to
disaggregate.

%% \begin{itemize}
%% \item Train on more data!
%%   \item Experiment with unsupervised pre-training on unlabelled
%%     aggregate power data.
%%   \item Combine all three approaches: pre-train a `rectangles' net on
%%     unlabelled data as an autoencoder.  Then attach an RNN to the
%%     output to capture detailed temporal patterns.  Or use an ensemble
%%     of multiple different approaches.
%%   \item Experiment with more permutations of the nets.
%%   \item Experiment with dropout and batch normalisation.
%%   \item Try training one large net to do multiple appliances.
%%   \item Improve 'rectangle' method to output multiple states per
%%     appliance.
%%   \item Try other input features:  time of day, day of week, season,
%%     temperature etc.
%%   \item Build more sophisticated synthesiser of aggregate data.
%%   \item Experiment with ways to allow give the network information
%%     about the absolute power (instead of independently centring each
%%     input sequence) whilst also allowing the network to generalise
%%     well.
%%   \item Try variational autoencoders.
%%   \item Generate a probabilistic output (either using existing `layering'
%%     approach or mixture density networks or variational approaches).
%%   \item Do fully integrated, multi-appliance disaggregation: use
%%     discrete optimisation to find most likely set of appliances.  Or
%%     an RNN which sees aggregate data as well as output of upstream
%%     appliance classifier.
%% \end{itemize}

%ACKNOWLEDGMENTS are optional
\section{Acknowledgments}
Jack Kelly's PhD is funded by the EPSRC and by Intel via their EU
Doctoral Student Fellowship Programme.  The authors would like to
thank Pedro Nascimento for his comments on a draft of this manuscript.

% The following two commands are all you need in the
% initial runs of your .tex file to
% produce the bibliography for the citations in your paper.
% \bibliographystyle{./abbrvurlmendeley.bst}
% \bibliography{library.bib}
% ACM needs 'a single self-contained file'!

\balancecolumns
\end{document}